\pdfoutput=1
\documentclass[10pt,twocolumn,letterpaper]{article}

\usepackage[pagenumbers]{wacv} 
\usepackage{siunitx}

\definecolor{wacvblue}{rgb}{0.21,0.49,0.74}
\usepackage[pagebackref,breaklinks,colorlinks,allcolors=wacvblue]{hyperref}

\title{KODAMA: Multimodal Digital Twin Reconstruction for Urban RF Propagation Modelling}

\author{
  Maximiliano Wardle$^{1,2}$\qquad A. Ryo Koblitz$^1$ \\
  $^1$Nokia Bell Labs \qquad $^2$University of York \\
  {\tt\small cgf520@york.ac.uk, ryo.koblitz@nokia-bell-labs.com}
}

\begin{document}
\twocolumn[{%
\renewcommand\twocolumn[1][]{#1}%
\maketitle
\begin{center}
  \includegraphics[width=\textwidth]{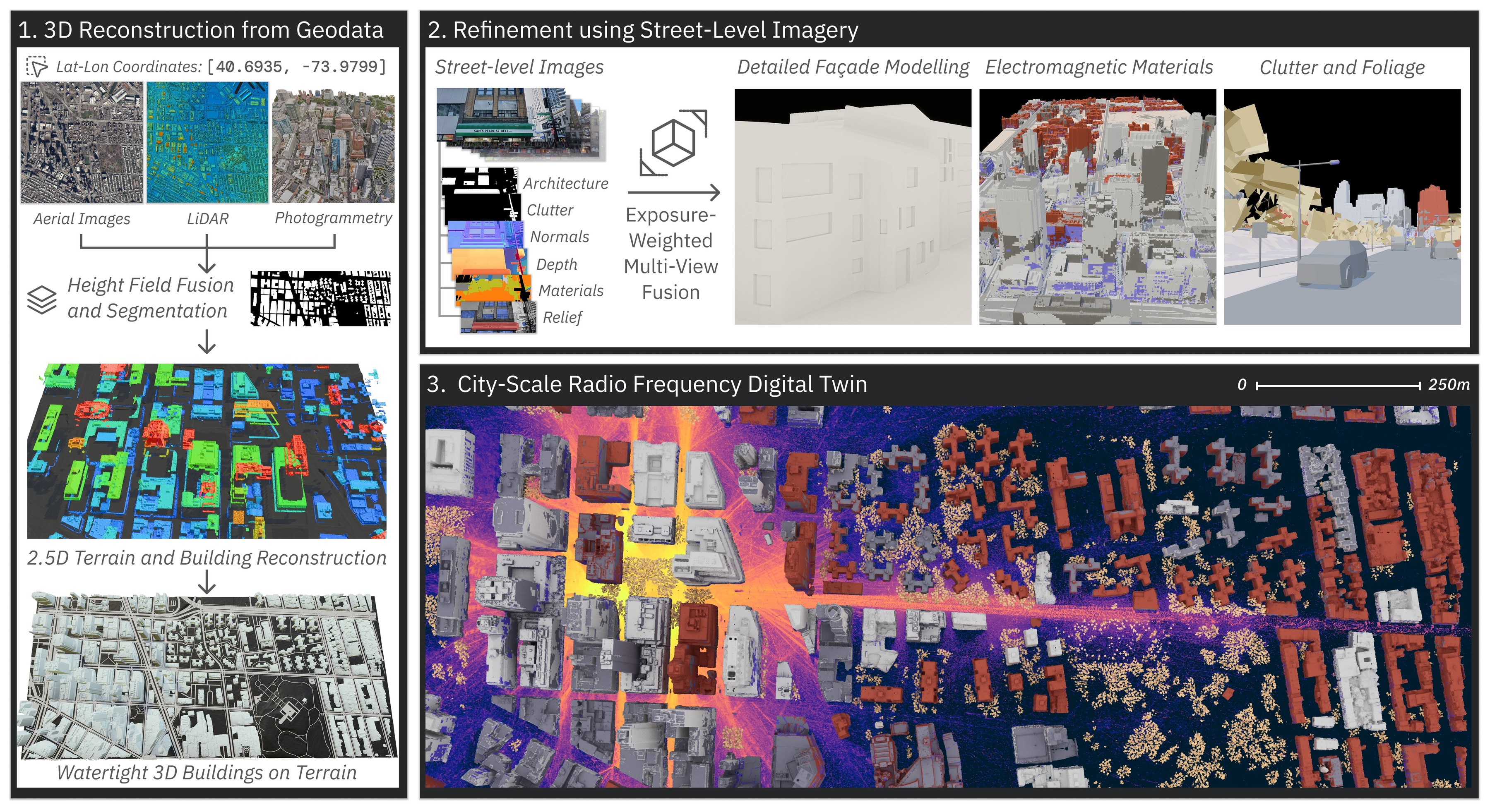}
  \captionof{figure}{KODAMA automatically builds site-specific RF digital twins from off-the-shelf geospatial data. 1. Aerial imagery, LiDAR, and photogrammetry are fused into 2.5D height fields yielding watertight buildings on bare-earth terrain. 2. Exposure-weighted multi-view fusion of street-level imagery recovers façade detail, per-surface electromagnetic materials, and clutter. 3. The twin predicts real-world measurements to single-digit RMSE without site visits or calibration; shown: radio coverage map over \qty{1.19}{\km^2} of Brooklyn.}
  \label{fig:teaser}
\end{center}}]
\renewcommand{\thefootnote}{\fnsymbol{footnote}}
\footnotetext[1]{Work completed during an internship at Nokia Bell Labs.}
\renewcommand{\thefootnote}{\arabic{footnote}}
\begin{abstract}
3D reconstruction typically strives for geometric fidelity or visual plausibility. Radio frequency digital twins (RFDT) are instead judged by whether communication channels behave in them as they do in the real world. RFDTs promise site-specific channel prediction but current practice forces a choice between coarse automated scenes and hand-built, measurement-calibrated models that take weeks to construct per-site. We present KODAMA, an automated pipeline that reconstructs ray tracing-ready RFDTs at city scale from off-the-shelf geospatial data alone: aerial imagery, LiDAR, and photogrammetry yield terrain and watertight building meshes, while exposure-weighted multi-view fusion of street-level imagery recovers fa\c{c}ade relief, electromagnetic materials, and clutter---all without site visits or calibration. Across three sites spanning 3.6 to 28\,GHz, KODAMA's uncalibrated predictions achieve single-digit RMSE, reducing point-to-point error by up to 5.35\,dB over automated baselines and coming within 0.22\,dB of a measurement-calibrated, hand-built RFDT.

\end{abstract}
    
\section{Introduction}
\label{sec:intro}

3D reconstruction and generation methods now produce increasingly realistic urban scenes at ever larger scale, supporting applications ranging from content creation to autonomous driving \cite{qian2026abotearth05generative3d, Liu2026-urbanverse, liu2025citygaussianv2efficientgeometricallyaccurate}. While these scenes are typically created and evaluated with visual plausibility and geometric fidelity in mind, they are increasingly asked to act as reliable stand-ins for real places in complex physics simulations \cite{Pa_en_2024, Jambon2026-wods, Fujiwara_2026, Zhu2026-gpr, Fang2025-scanbimsim}. Radio frequency  (RF) propagation modelling is one such domain.

Given a 3D representation of a real site---a geospatial digital twin---augmented with the electromagnetic properties of its surfaces, communication channels between transmitters and receivers can be generated via a suitable RF propagation model (here, ray tracing-based). We call this combination of scene and propagation model an RF digital twin (RFDT). 
RFDTs are positioned as sources of site-specific radio channels consumed by network tasks such as coverage prediction, infrastructure planning, resource allocation and spectrum management~\cite{alkhateebRealTimeDigitalTwins2023}, and are proposed as a source of training data for signal processing algorithms that currently govern network behaviour \cite{Honkala_2021, Agheli_2025, Ait_Aoudia_2022, Nachmani_2018}. 
The performance of these models hinges on the fidelity of the twins they are trained on~\cite{luo2025digital,ruahHowBridgeSimtoReal2026} and errors in a scene surface not as visual artifacts but as degraded decisions in the networks.

Crucially, this fidelity can be measured. An RFDT is judged not by resemblance to reference geometry but by how closely its simulated channels match measurements recorded at the site. A scene may therefore score well on conventional reconstruction metrics and still be wrong in ways that matter for propagation.

In this work, we demonstrate that predictive, site-specific RFDTs matching ground-truth signal strength measurement with single digit root mean squared error (RMSE) can be generated automatically, at city scale, from geospatial data that already exist. 
Current methods for building wireless digital twins present a trade-off between scale, fidelity and acquisition effort.
To our knowledge, KODAMA is the first demonstrably generalisable RFDT to produce state-of-the-art channel predictions without measurement or site calibration. It takes off-the-shelf aerial imagery, photogrammetry and LiDAR products, extracts terrain, roads and building volumes, then uses street-level imagery to assign dielectric materials, recover façade relief, and place clutter such as foliage, street furniture and vehicles (Fig. ~\ref{fig:teaser}). High-fidelity, kilometre-scale, ray tracing-ready scenes can be produced in hours instead of weeks.

Our main contributions are:
\begin{itemize}
\item Desiderata for RF-optimised scene reconstruction — what ray tracing requires that geometric and perceptual measures do not guarantee.
\item KODAMA, an automated pipeline that fulfils these requirements, using widely available spatial data to reconstruct high-fidelity ray tracing-ready RFDTs of real world sites at city scale, without the need for site visits, bespoke capture, or manual modelling.
\item Validation against real-world measurements at three sites, showing that KODAMA substantially improves on existing automated approaches to reconstructing RFDTs, and that its uncalibrated predictions rival hand-built twins calibrated against on-site measurements.
\end{itemize}

\section{Radio-frequency ray tracing desiderata}
\label{sec:desiderata}
Ray tracing may be best known as a 3D rendering technique, but it is also a potent tool for RF propagation modelling ~\cite{yunRayTracingRadio2015,fuschiniRayTracingPropagation2015}. Light and radio are both electromagnetic waves, but radio operates at wavelengths orders of magnitude larger. This not only necessitates a higher-order approximation to Maxwell's equations---it fundamentally reorients scene reconstruction from generating plausible views to predicting RF propagation. The level of detail scenes need is also not fixed: as wireless networks move to higher frequencies and shorter wavelengths, smaller features have a greater effect on propagation.

The following desiderata make these requirements explicit, identifying scene properties that matter for RF ray tracing but are not always guaranteed by the conventional methods and metrics used in 3D scene reconstruction:

\begin{itemize}
    \item \textit{Watertight polygonal meshes.} RF ray tracers require unambiguous intersections with explicit surfaces \cite{hoydis2023sionnart}. Holes, self-intersections and non-manifold edges can let rays pass through otherwise solid objects or create spurious reflections. Each physical object should therefore form a closed, manifold mesh.
    \item \textit{Semantic separability.} Objects interact with signals differently and must remain separable---both to receive correct treatment and to attribute channel effects to their causes.
    \item \textit{Accurate terrain topography.} Terrain sets the datum for every object placed on it, determining line of sight and how signals interact with buildings and the ground~\cite{Chizhik_2010,Chizhik_2020}. 
    \item \textit{Correct building volumes.} Buildings are the dominant blockers and reflectors of RF; errors in footprint, height or roof shape shift shadow boundaries and create, remove or misplace propagation paths.
    \item \textit{Detailed façades.} At higher frequencies, façade details such as balconies, window frames and roof ledges go from negligible surface texture to significant blockers or scatterers. RFDTs therefore require detailed façades grounded in real observations rather than invented geometry.
    \item \textit{Per-surface electromagnetic materials.} Material electromagnetic properties determine reflection strength and how much energy is lost with each bounce \cite{itur2023p2040}. To model these behaviours, RF ray tracing requires relative permittivity and conductivity rather than the bidirectional reflectance distribution function (BRDF) parameters typically used to model surface appearance. 
    \item \textit{Clutter.} `Clutter' objects such as vegetation, street furniture, signage and cars are not merely decorative: they block, weaken or redirect signals \cite{heijsImportanceScatteringPoles2024,Chizhik_2022, gomez-ponceEfficientIntegrationStreet2026}.
    \item \textit{Georeferencing.} Objects must be positioned correctly relative to each other and in real-world coordinates, so that diverse data fuse cleanly and predictions transfer to the site.
\end{itemize}

\section{Related work}
Existing approaches to scene reconstruction satisfy different subsets of the desiderata laid out above, yet at present, no automated method exists for reconstructing city-scale 3D environments for RF ray tracing that fulfils them all.
\subsection{RF digital twin reconstruction}

Typical automated pipelines for building RFDTs such as OpenGERT~\cite{tadikOpenGERTOpenSource2025}, Geo2SigMap~\cite{li2024geo2sigmaphighfidelityrfsignal} and OpenPathNet~\cite{liu2026openpathnetopensourcerfmultipath} extrude 2D OpenStreetMap (OSM)~\cite{OpenStreetMap} footprints to estimated heights on SRTM~\cite{NASA_SRTMGL1_003} terrain. SceneBaker~\cite{lyuSceneBakerRadioReadyScene2026} adds LLM-driven material assignment, though without validation against measurements. None recover façade detail or clutter, and their fidelity is bounded by their inputs: SRTM's circa 30 m resolution and 16 m vertical accuracy \cite{NASA_SRTMGL1_003} are too coarse for urban RF modelling, while OSM buildings are frequently misshapen, displaced or missing, with less than \qty{8}{\%} carrying height data~\cite{Fan_2014,Biljecki_2023}. Resulting scenes reduce buildings to flat-topped prisms extruded to approximate heights, and simulations based on them depart significantly from real-world measurements.

At the opposite extreme of fidelity, highly-detailed RFDTs are reconstructed for individual sites by hand. Ying et al.~\cite{ying2025sitespecific} survey a nearly 2 km² area in downtown Brooklyn using rangefinder measurements, LiDAR and photogrammetry, then use this data to manually refine OSM geometry, correcting building dimensions, modelling façade details, assigning materials and placing clutter and foliage. The resulting twin matches real-world measurements to within 4 dB, but is highly labour-intensive to create. In Sec.~\ref{sec:evaluation}, we use KODAMA to reconstruct the same site automatically from open geospatial data and existing street-level imagery.

Others modify open CityGML models, LiDAR or photogrammetry released by local or national governments for use in RF ray tracing~\cite{testolinaBostonTwinBoston2024,iyeOpenWirelessDigital2025}. These datasets may contain richer geometry than OSM-derived scenes, but they offer models of individual cities rather than a generalisable method for automatically reconstructing RFDTs.

Emerging vision-based methods automate high-fidelity reconstruction for RF ray tracing: HoRAMA~\cite{ying2026horama} builds twins from smartphone video and uses a vision-language model to assign materials, matching measurements to within \qty{2.28}{\dB}, while RFDT-Channel~\cite{yao2026rfdt}, VisRFTwin~\cite{an2026tamingvisionpriorsdata} and RadarTwin~\cite{bejerano2026radartwinscenespecificmmwaveradar} reconstruct from multi-view imagery and/or LiDAR. NeRF\textsuperscript{2}~\cite{zhaoNeRF2NeuralRadioFrequency2023}, RF-3DGS~\cite{zhangRF3DGSWirelessChannel2025} and WRF-GS~\cite{wen2025wrfgs} instead learn neural radio fields directly from RF measurements. All, however, require dedicated site-specific capture and have been demonstrated only in small, bounded, mostly indoor environments. KODAMA instead builds complete outdoor twins at city scale from pre-existing geospatial data, with no site-specific capture or calibration.

\noindent\subsection{Urban scene reconstruction beyond RF}
Current methods for building wireless digital twins  are limited in scale, in fidelity, or in the amount of capture they demand: none produces high-fidelity city-scale scenes automatically from pre-existing data alone. We therefore look beyond RF to the broader state of the art in urban reconstruction and physical simulation.

Recent neural methods~\cite{Gao_2025,chen2024gigagsscalingplanarbased3d, liu2025citygaussianv2efficientgeometricallyaccurate, Li_2025, chen2026metrogsefficientstablereconstruction} reconstruct detailed city-scale scenes from large collections of images. Beyond requiring extensive site-specific multi-view capture, these approaches optimise representations for photorealism or generic geometric accuracy rather than physical simulation and are not designed to provide semantically separated, regularised and watertight buildings and terrain. 

ABot-Earth~\cite{qian2026abotearth05generative3d}, Sat2City v2~\cite{hua2026sat2cityv2native3d}, Sat2RealCity~\cite{wang2026sat2realcitygeometryawareappearancecontrollable3d}, EarthCrafter~\cite{Liu2026-ze} and related methods require substantially less input data, generating large urban environments from satellite imagery. While their outputs look plausible, they are not faithful twins of real-world sites. Building footprints and roofs may be constrained to real data, but vertical or occluded surfaces such as façades are synthesised from learned priors. In RF ray tracing, inventing detail is worse than simply omitting it, as it produces propagation paths that do not exist in the real site. KODAMA instead adds detail supported by real observations and retains the evidence supporting each geometric operation.

Next, we look to explicit reconstruction techniques that create meshes from geospatial survey data and emphasise geometric accuracy and topological validity. Such methods are favoured in adjacent physical simulation domains such as computational fluid dynamics (CFD) which, like RF ray tracing, require accurate, watertight geometry.

Direct 3D methods reconstruct building meshes from point clouds~\cite{Chen_2024,Chen_2022,Bauchet_2020}, while 2.5D approaches instead take advantage of widely available raster data including orthophotography, height maps and LiDAR point clouds~\cite{Schuegraf_2024, Xu_2024, Schuegraf_2025, Abdelhedi_2026, Yi_2026}. Despite their differing inputs, both generally use piecewise-planar primitive assembly to produce compact, regular models with clean rooflines, but this comes at the cost of curved and irregular features like domes or rounded walls. Moreover, because their inputs are predominantly captured from overhead, façades are weakly observed and therefore remain featureless vertical faces in the resulting models. 

Paden et al.~\cite{Pa_en_2024} use a 2.5D approach to automatically reconstruct large urban areas for CFD from open LiDAR data, but identify the absence of façade detail as a key limitation. Their workflow also omits per-surface materials and clutter. 
Closest to our approach, VoxCity~\cite{Fujiwara_2026} automatically fuses open geospatial data into voxel models of almost any city for large-scale simulation, but its geometric resolution is too coarse for RF ray tracing at the frequencies we consider, and despite an initial trial of street-image refinement it does not yet resolve detailed building geometry, per-surface materials or clutter.

Methods such as~\cite{Wysocki_2023, Hanke_2025, Tang_2025, Ma2026-yk} aim to close this gap. They refine existing models by adding features such as doors and windows using measurements taken from mobile laser scans or street-level imagery. However, these require accurate, well-structured geometric priors which may or may not exist for the target site.
\section{Method}
KODAMA reconstructs buildings as watertight solids directly from dense height fields, preserving complex building and roof shapes, then uses street imagery to recover façade geometry, assign electromagnetic materials and populate the scene with clutter.

Our approach is deliberately source-agnostic, allowing KODAMA to use whichever datasets provide the best coverage of a target site. It accepts either:
\begin{itemize}
    \item A georeferenced orthophoto together with a LiDAR point cloud, raster Digital Surface Model (DSM), or both. Adding a bare-earth Digital Elevation Model (DEM) improves terrain quality but is not strictly required. KODAMA automatically aligns these datasets within a common geospatial frame.
    \item Georeferenced photogrammetric models, usually from an OGC 3D Tiles-compatible source ~\cite{Cozzi2023-3t}.
\end{itemize}

KODAMA converts the supplied data into a shared set of 2.5D height fields, in which each horizontal location $\mathbf{x}=(x,y)$ on a shared georeferenced grid $\Omega$ stores a single surface elevation $z$. Each source $s$ contributes one such field, $H_s:\Omega\rightarrow\mathbb{R}\cup\{\bot\}$, taking the value $\bot$ wherever that source observed nothing, and we write $H$ for their composite. This recasts many complex 3D reconstruction tasks as simpler image-processing problems, and allows the same set of operations to be applied regardless of the number or type of sources used. Although this intermediate representation does not encode multiple surfaces at a single horizontal location, KODAMA recovers these structures in subsequent stages.

\subsection{Reconstructing terrain and buildings}
\label{sec:terrain-buildings}
Where a high-resolution DEM is available for the site, we use it directly as the scene terrain. Otherwise, KODAMA estimates the terrain surface through iterative low-elevation sampling of mesh-ray intersections or LiDAR returns, beginning with a coarse, conservative ground estimate and progressively increasing the sampling resolution.

At pass $k$, $\mathcal{Z}_k(c)$ contains elevations returned by jittered rays in coarse cell $c$ after discarding hits whose vertical normal component is below $n_{\min}$. Its $q_k$-th percentile, $\bar T_k(c)=\operatorname{perc}_{q_k}\mathcal{Z}_k(c)$, provides a
conservative estimate of local ground level. The acceptance mask $\chi_k(\mathbf{x})=\mathbf{1}[h_k(\mathbf{x})\leq \mathcal U(\bar T_k)(\mathbf{x})+\tau_k]$ retains a fine-grid hit $h_k(\mathbf{x})$ only where one is returned and it lies within a tolerance $\tau_k$ of the interpolated estimate $\mathcal{U}(\bar T_k)$; elsewhere we fall back on the estimate itself and gently smooth the result with a Gaussian $G_{\sigma_k}$:
\begin{equation}
T_k=G_{\sigma_k}*
\left[
\chi_kh_k+(1-\chi_k)\mathcal U(\bar T_k)
\right].
\label{eq:terrain}
\end{equation}
As the resolution increases, tightening $\tau_k$ enables us to recover finer terrain variation without accidentally absorbing elevated structures like buildings or foliage. The final pass yields the terrain $T$, and subtracting it from the original height field gives the above-ground field $A(\mathbf{x})=\max(H(\mathbf{x})-T(\mathbf{x}),0)$, which we use to drive subsequent reconstruction stages.

Instead of relying on building footprints from OSM or other databases, we apply Girard et al.’s architecture~\cite{Girard_2021} to an orthographic view of the site to obtain building masks $F_b\subset\Omega$ with coherent boundaries, whilst setting unassigned above-ground regions aside so we can extract foliage from them later (see Sec.~\ref{sec:clutter}). Rather than extrude these footprints up to approximate heights, we mask the above-ground field with each building footprint to recover its roof surface in situ, $Z_b=A|_{F_b}$. We then identify and remove probable non-roof geometry using semantic labels where the source data provides them and geometric cues (chiefly how many surfaces are stacked vertically at a location, component size and connectivity) where it does not, then fill the resulting gaps to get the cleaned field $\widetilde Z_b$. Extruding each cleaned roof surface down to the bare-earth terrain gives the building solid $\mathcal{B}_b=\{(\mathbf{x},z):\mathbf{x}\in F_b,\;0\leq z-T(\mathbf{x})\leq\widetilde Z_b(\mathbf{x})\}$ 
whose boundary $\widehat B_b=\partial\mathcal{B}_b$ we remesh into one watertight polygonal mesh per building, welding disconnected roof regions into a single solid. This approach preserves each building’s footprint and roof shape---including curvilinear geometries and propagation-relevant features such as chimneys, HVAC equipment and solar panels---without the need for manual modelling.

\begin{figure*}[t]
  \centering
  \includegraphics[width=\textwidth]{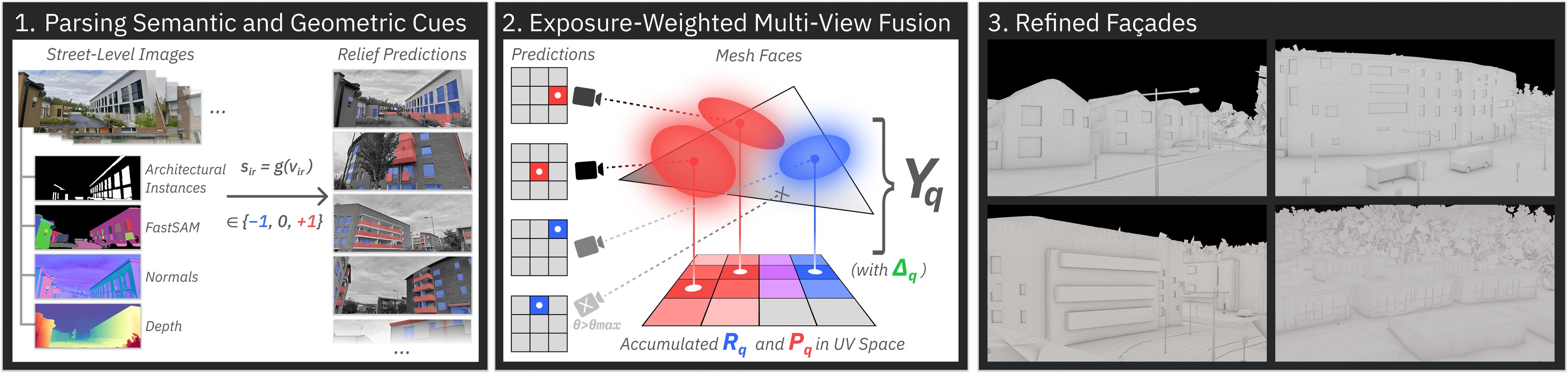}
  \caption{Fa\c{c}ade relief recovery. Per-region relief predictions from semantic and geometric cues (1) are projected onto surfaces by exposure-weighted multi-view fusion; accumulated evidence plus source-mesh discrepancy give a per-texel displacement posterior (2), discretised and applied by Boolean operations to produce relief absent from the source geometry (3).}
  \label{fig:fusion}
\end{figure*}

\subsection{Recovering detail from street-level imagery}
\label{sec:street-detail}
At this stage, KODAMA produces scenes that satisfy most of the desiderata for RF ray tracing, but three remain unmet: detailed façade geometry, electromagnetic materials and clutter elements such as street furniture and vegetation. All three lie precisely where the data sources used so far provide the least information: ground level.

To complete scenes, we turn to street-level imagery from crowdsourced platforms such as Mapillary~\cite{mapillary}, registering thousands of views per target area directly from provider metadata. Classical pose refinement is unreliable here because metadata quality varies widely across providers, and the landmarks it would anchor to are exactly the features still missing from the scene. We therefore adopt an exposure-aware multi-view fusion approach.

\subsubsection{Parsing street-level imagery}
\label{sec:parsing}
Having registered each street-level image in the scene, we use SAM3~\cite{carion2026sam3segmentconcepts} to generate two sets of masks deployed throughout the rest of the reconstruction process. This keeps every layer of the scene aligned while avoiding redundant segmentation.
\textit{Architectural instances} capture key architectural features relevant to RF propagation, including recessed windows and doors and protruding drainpipes and balconies. We use these architectural instances to guide façade geometry reconstruction in Sec.~\ref{sec:facade} and electromagnetic material assignment in Sec.~\ref{sec:materials}. 
\textit{Clutter instances} isolate common objects found in the urban environment such as foliage, street furniture and vehicles. We use these masks first to stop foreground clutter from contributing erroneous geometry or materials to building surfaces and then to populate the scene with 3D clutter objects.

\subsubsection{Reconstructing façade geometry}
\label{sec:facade}
To recover propagation-relevant architectural features currently absent from our building meshes, we start by inferring local façade relief in each street-level image. We use FastSAM~\cite{zhao2023fastsegment} to construct an overlap-free hierarchy of façade regions, then use the architectural instances to refine them, retaining only regions enclosed by a dominant wall. For each retained region, we combine semantic and geometric cues to estimate a signed relief field and classify it as protruding, recessed or neutral relative to its surroundings.

The architectural instances cover most elements of interest, but not all. We recover additional features semantic segmentation might otherwise miss by comparing each region’s estimated normal orientation (provided by RoSE~\cite{li2026monocularnormalestimationshading}) and relative depth (from DepthPro~\cite{bochkovskii2025depthprosharpmonocular}) with those of its neighbours. We also evaluate an image-plane flow field derived from both cues, providing additional directional evidence of whether a region projects from or recedes into its surroundings. To resolve ambiguous cases—for example, a window frame that may be recessed into a wall while still projecting beyond the glass panes it surrounds—we define and apply an additional set of coherence rules. We summarise this process for region $r$ in image $i$ as $\mathbf{v}_{ir}=(v^{\mathrm{normal}}_{ir},v^{\mathrm{depth}}_{ir},v^{\mathrm{flow}}_{ir},v^{\mathrm{semantic}}_{ir})$ and $s_{ir}=g(\mathbf{v}_{ir})\in\{-1,0,+1\}$, denoting recessed, neutral and protruding areas.

Next, we project these per-image predictions onto the scene by exposure-weighted multi-view fusion, the operator we reuse for material assignment in Sec.~\ref{sec:materials}. We cast rays from every street image pixel through world space and, at each hit on a building surface, splat a signed Gaussian kernel into UV space~\cite{Zwicker_2001}, where the kernels accumulate as votes that the surface should move inward or outward.

For pixel $p$ in image $i$, let $\mathbf{u}_{ip}=\pi_{\mathrm{UV}}(\mathbf{x}_{ip})$ be its surface hit in UV space, and let $\theta_{ip}$ be the angle between the surface normal $\mathbf{n}_{ip}$ at that hit and the viewing direction $\mathbf{d}_{ip}$, so that $\cos\theta_{ip}=-\mathbf{n}_{ip}^\top\mathbf{d}_{ip}$. For focal length $f_i$ and hit distance $t_{ip}$, the projection Jacobian $J_{ip}=\cos\theta_{ip}\,(f_i/t_{ip})^2$ counts, up to a constant, how many pixels of image $i$ fall on a unit of surface area there. We gate it to zero for views more oblique than $\theta_{\max}$. Weighting each kernel by the projection Jacobian therefore prefers close, head-on views to distant or grazing ones. For stage $a\in\{\mathrm{fac},\mathrm{mat}\}$, \begin{equation}
\omega^{(a)}_{ipq}
=J_{ip}^{\gamma_a}K_\sigma(q-\mathbf{u}_{ip}),
\qquad
W_{\mathrm{exp}}^{(a)}(q)=\sum_{ip}\omega^{(a)}_{ipq},
\label{eq:projection}
\end{equation}
where the stage exponent $\gamma_a$ sets how strongly that preference is applied. We use a high $\gamma_{\mathrm{fac}}$ for façade relief, where geometric accuracy is key, and a lower $\gamma_{\mathrm{mat}}$ for materials (Sec.~\ref{sec:materials}), where we aim for maximum coverage. The façade exposure map $W_{\mathrm{exp}}^{(\mathrm{fac})}$ records how confidently the available views observe each part of the surface, so that texels with little exposure are treated as unobserved rather than as evidence for a flat surface. 
Separated by class, the inward and outward votes give totals $E^{\pm}_q=\sum_{ip}\omega^{(\mathrm{fac})}_{ipq}\,\mathbf{1}[s_{ip}=\pm1]$ where $s_{ip}$ is the class of the region containing pixel $p$. Normalising by $W_{\mathrm{exp}}^{(\mathrm{fac})}(q)$ and discretising gives the recessed and protruding evidence states $R_q$ and $P_q$, so a texel seen by many views is not favoured over one seen by few. For buildings derived from photogrammetry, we add the signed discrepancy $\Delta_q$ between the reconstructed façade and source geometry as a third signal.

Finally, we combine these signals into a displacement class by discrete Bayesian fusion with an expert-specified conditional probability table. Together they form the evidence tuple $\mathbf e_q=(R_q,P_q,\Delta_q)$. The discrepancy does not enter directly, but rather induces a soft distribution $\Pr(d\mid\Delta_q)$ over the latent mesh-discrepancy state $d$, which is neutral when no source mesh is available. Summing over $d$ gives the posterior over displacement class $Y_q$ at texel $q$:
\begin{equation}
\Pr(Y_q=y\mid\mathbf e_q)
=\sum_d
\Pr(Y_q=y\mid d,R_q,P_q)\Pr(d\mid\Delta_q).
\label{eq:facade-fusion}
\end{equation}
Treating the source mesh as evidence about $d$ rather than as a direct observation lets the photogrammetric geometry modulate the image evidence instead of overriding it, so a confident, well-observed relief prediction can outweigh a base mesh that disagrees. The posteriors produced are still soft fields, so we convert them into discrete instances using thresholding, morphological filtering, and connected-component analysis, enabling us to drop weaker predictions, suppress noise, close small gaps and separate merged regions. Each surviving candidate is then mapped back to the building surface in world space and converted into a 3D prism used to apply Boolean operations, subtracting recessed features from the façade and adding protruding ones to yield a final detailed building mesh.
\begin{figure*}[t]
  \centering
  \includegraphics[width=\textwidth]{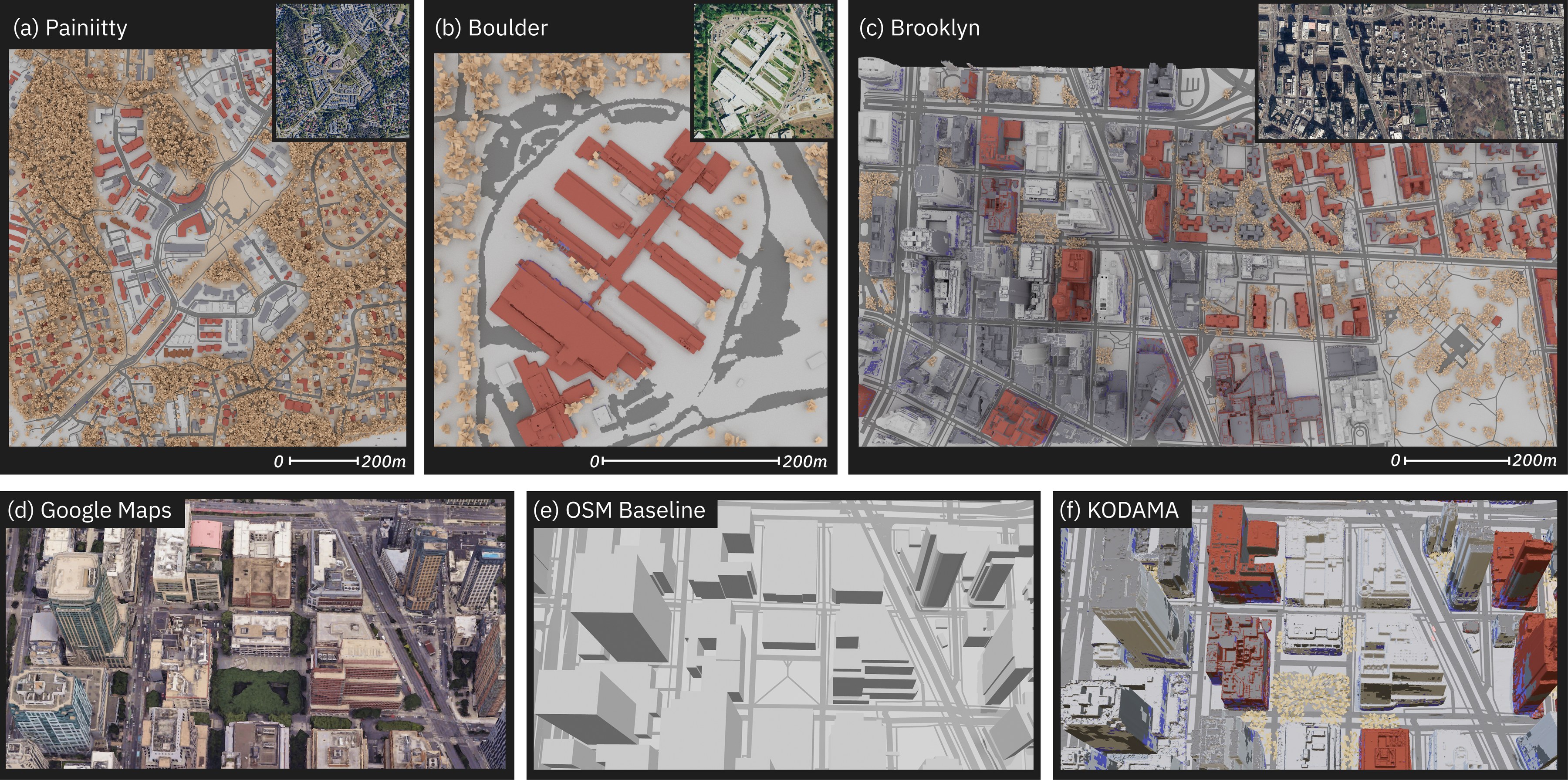}
  \caption{Top: KODAMA reconstructions of the three evaluation sites —(a) Painiitty, suburban with extensive vegetation; (b) Boulder, a single large laboratory building; (c) Brooklyn, dense urban canyon. Insets show aerial imagery of the same area. Bottom: a detail view of one block in Brooklyn as seen from Google Maps, Imagery ©2026 Google (d), the automated OSM baseline (e) and KODAMA (f). OSM reduces buildings to bare flat-topped prisms with no vegetation or clutter; KODAMA recovers roof structure, fa\c{c}ade relief, clutter and per-surface materials (false-coloured by ITU-R P.2040 class).}
  \label{fig:results}
\end{figure*}
\subsubsection{Assigning electromagnetic materials.}
\label{sec:materials}
Whereas photorealistic scene reconstruction strives for accurate appearance and sharp material boundaries, we seek to recover the most likely spatial distribution of electromagnetic properties across every surface. Our approach therefore favours broad material coverage, even with approximate boundaries, over more conservative methods that might omit key material regions altogether.

We begin by applying RMSNet~\cite{Cai_2024, Huynh2025-lq} to predict one of 20 material classes for each pixel in every street-level image, then map these classes to the material models in ITU-R P.2040-4~\cite{itur2023p2040}, a standardised set of relative permittivity and conductivity parameters for common building materials that is widely used in RF ray tracing. 

Boundaries between materials are particularly important around electromagnetically significant features such as windows and doors, so we refine RMSNet’s predictions using the architectural instances, selecting the dominant compatible material class for each instance and applying it throughout the mask. We then use the clutter instances to prevent foreground objects’ materials being erroneously transferred onto building façades.

To assign these materials to the scene geometry, we reuse the exposure-weighted multi-view fusion of Sec.~\ref{sec:facade} with exponent $\gamma_{\mathrm{mat}}$, but each pixel now votes for a material class rather than a relief class. Writing $m_i(p)$ for the material predicted at pixel $p$ in image $i$, and $\mu_m$ for a class-specific vote weight, we select at each texel $q$
\begin{equation}
\widehat m(q) = \arg\max_m \;\mu_m
\sum_{ip} \omega^{(\mathrm{mat})}_{ipq}
\mathbf{1}[m_i(p)=m].
\label{eq:material-fusion}
\end{equation}
Agreement across views therefore strengthens a class, while conflicting observations compete in proportion to their projection weights. The selected class is mapped to its electromagnetic parameters as $(\varepsilon_r(q),\sigma_{\mathrm{EM}}(q))=\lambda(\widehat m(q))$.

If required, we extend the dominant material onto unobserved areas of the same surface---for example, from the visible portion of a roof to the remainder---while retaining hit counts and $W_{\mathrm{exp}}^{(\mathrm{mat})}$ to distinguish direct evidence from extrapolated assignments. The result is comprehensive RF material coverage across every object in the scene.

\subsubsection{Placing RF clutter and foliage.}
\label{sec:clutter}
We recover clutter objects by taking each instance and estimating its metric depth in the image using DepthPro. We then calibrate these estimates against geometry already in the scene, using sparse ray casting to obtain a set $C_i$ of pixels with both a predicted depth $d_i(p)$ and a scene-ray distance $t^{\mathrm{scene}}_i(p)$. We fit a robust affine calibration $(\widehat{a}_i,\widehat{b}_i)
=\operatorname{TheilSen}\bigl\{(d_i(p),t^{\mathrm{scene}}_i(p))\bigr\}_{p\in C_i}$ 
and use $\widehat{d}_i(p)=\widehat{a}_i d_i(p)+\widehat{b}_i$ when lifting instances into world space. We then reconcile repeated detections through spatial clustering and non-maximum suppression, apply a simple scene grammar to correct implausible placements, and instantiate a class-specific mesh from a library of pre-authored assets at each surviving location. This produces a diverse, spatially grounded distribution of clutter without needing to reconstruct each individual object directly from street-level imagery.

Foliage requires a different treatment. Individual smaller vegetation instances are identified and placed like other clutter objects. For larger tree canopies, we return to the above-ground regions set aside in Sec.~\ref{sec:terrain-buildings}, keep those that are large, elevated and geometrically irregular, and group connected residuals into contiguous canopy envelopes. Following~\cite{chizhikUniversalPathGain2022, RECOMMENDATIONITURP83310} and~\cite{gomez-ponceEfficientIntegrationStreet2026}, we represent these envelopes as absorbing volumes with cumulative loss $L_{\mathrm{fol}}(\rho,\nu)=\alpha_\nu\sum_k\ell(\rho\cap V_k),$ for a ray $\rho$ at band $\nu$ 
where $\ell(\rho\cap V_k)$ is the path length through canopy volume $V_k$ and $\alpha_\nu$ the band-specific attenuation per unit distance. Within each volume, we also procedurally distribute planar meshes to provide explicit surfaces for reflection and scattering. Together, these give a hybrid volumetric--polygonal representation that captures the principal effects of foliage on signal propagation while remaining tractable at city scale.

\section{Evaluation and results}
\label{sec:evaluation}

We evaluate KODAMA’s performance by reconstructing three sites spanning different built environments and input data modalities, then compare ray traced predictions of RF propagation in each scene against independent real-world measurements captured at the same location. 

At each site, we benchmark KODAMA against a baseline OSM scene generated using the Blosm plugin~\cite{Elistratov2014-bl} from SRTM terrain and OpenStreetMap buildings, following the same construction used by prior automated geospatial pipelines~\cite{tadikOpenGERTOpenSource2025,li2024geo2sigmaphighfidelityrfsignal,liu2026openpathnetopensourcerfmultipath}. Such scenes carry no façade relief, clutter or vegetation. 
We also compare against 3GPP TR 38.901~\cite{3gpp38901}, the industry-standard statistical model, which consumes no 3D scene. Rather, channels are drawn from probabilistic rules conditioned on coarse scenario labels and transmitter–receiver distance.
For Brooklyn, we add Ying et al.'s~\cite{ying2025sitespecific} published results for their  NYURay twin, enabling us to compare KODAMA's automated, uncalibrated predictions to a hand-built model calibrated against measurement data.

Radio propagation is modelled using NVIDIA Sionna RT~\cite{hoydis2023sionnart}. Paths are computed with line-of-sight, specular and diffuse reflection, refraction, and diffraction enabled, and these paths are traced with between 12 and 20 allowable interactions with the environment. Where foliage is present, the same hybrid vegetation construction is used: discrete canopy scatterers remain in the tracer, while bulk crown occupancy is applied as a post-trace attenuation along the path. Solver settings are held fixed for each scene so that differences in predicted channels can be attributed to the reconstruction rather than to the RF ray tracer.

\begin{table}[t]
\centering
\caption{Trend-line and point-to-point RMSE versus measurements.
PD is compared with 3GPP TR~38.901.
CI is compared with NYURay. For all three metrics, lower is better.}
\label{tab:scene-rmse}
\begin{tabular}{llcccc}
\toprule
Site & Metric & \multicolumn{4}{c}{RMSE (dB) $\downarrow$} \\
\cmidrule(lr){3-6}
 & & Ours & OSM & 3GPP & NYURay \\
\midrule
Painiitty & P2P& \bf{7.76} & 13.11 & --- & ---\\
Painiitty & PD & \bf{4.09} & 6.75 & 15.61 & --- \\
Boulder   & P2P& 1.99 & \bf{1.84} & --- & --- \\
Boulder   & PD & 1.63 & \bf{1.51} & 5.90 & --- \\
Brooklyn  & P2P& \bf{7.09} & 9.05 & --- & --- \\
Brooklyn  & PD & \bf{3.96} & 5.03 & 10.04 & --- \\
Brooklyn  & CI & 3.92 & 4.81 & --- & \bf{3.70} \\
\bottomrule
\end{tabular}
\end{table}
RFDT evaluations typically centre on trend-line scores (power-distance, PD, or close-in, CI) that coarsely quantify how radio signal loss grows with distance. A stricter test is point-to-point (P2P) transmitter-receiver pair comparisons. P2P evaluations reveal local, scene specific effects, e.g.~`around the corner' propagation due to street clutter diffraction~\cite{Chizhik_2022}, that are masked in trend-line metrics like PD and CI, but are less frequently reported in the literature.

\noindent\textbf{Painiitty.} Our first study area covers \qty{1.54}{\km^2} of suburban Espoo, Finland, where low-rise buildings sit between several large wooded areas. We use photogrammetry from the Helsinki Map Service \cite{helsinki3dmesh2017} and street-level imagery from Mapillary \cite{mapillary} as inputs to reconstruction. The scene is evaluated using measurements from an industry drive-test campaign at \qty{3.6}{\GHz}. In both P2P and PD KODAMA outperforms the OSM and 3GPP baselines by a wide margin, as shown in table~\ref{tab:scene-rmse}. The large gap between KODAMA and the 3GPP baseline highlights the inability of trend-line-based modelling to capture strong local variations encountered in realistic deployment scenarios. Against the base OSM scene, KODAMA cuts P2P RMSE by \qty{5.35}{\dB} and PD RMSE by \qty{2.66}{\dB}. As dB is a log-power scale, reducing P2P RMSE from \qty{13.11}{\dB} to \qty{7.76}{\dB} is a 3.4$\times$ reduction in typical multiplicative power error, not a \qty{41}{\%} improvement. This gap arises primarily in non-line of sight (NLOS) conditions due to the lack of vegetation and incorrect building volumes in the OSM scene.

\noindent\textbf{Boulder.} The second site we consider spans \qtyproduct{500 x 500}{\m} around the National Institute of Standards and Technology (NIST) campus in Boulder, Colorado, and is dominated by a single sprawling laboratory building measuring more than \qty{300}{\m} across. Scene geometry is derived from aerial LiDAR \cite{DRCOG2022-rl} and NAIP orthoimagery \cite{NAIP2021-CO} obtained via the Colorado Spatial Data Portal, then refined using panoramic image sequences recorded during a NIST measurement campaign. We evaluate the reconstruction against 4,979 measurements collected along densely sampled robot trajectories in two outdoor areas, a courtyard and a walkway at a frequency of \qty{28}{\GHz}~\cite{nist_nextg}. Here, KODAMA and OSM are closely matched (within around \qty{0.1}{\dB}), both outperforming 3GPP PD predictions. Because the measured transmitter--receiver pairs are almost entirely line-of-sight, predicted loss is dominated by the direct path regardless of how the surrounding scene is modelled.

\noindent\textbf{Brooklyn.} Our final scene takes in approximately \qtyproduct{1.7 x 0.7}{\km} around MetroTech Commons and along Myrtle Avenue in Brooklyn, New York. This dense downtown urban area contains numerous high-rise towers, several over \qty{100}{\m} tall. Reconstruction uses open LiDAR point clouds, a DEM, \cite{NYC2017-topobathy} and orthophotos~\cite{NYS2024-ortho} from New York State's GIS Clearinghouse, together with Mapillary street imagery. We evaluate the scene with measurements from twenty fixed transmitter–receiver pairs at \qtylist{6.75;16.95}{\GHz}~\cite{shakyaUrbanOutdoorPropagation2025}. The hand-built, measurement-calibrated NYURay twin remains ahead of KODAMA by just \qty{0.22}{\dB} in CI. KODAMA significantly outperforms 3GPP in PD, and beats OSM in all three measures.

\section{Limitations and future work}
KODAMA generates scenes automatically from off-the-shelf geodata and therefore depends on what exists for the target site; coverage of these products is uneven and concentrated in urban centres and the Global North. The 2.5D height-field representation lets KODAMA draw on many data sources, with fine-grained relief added from street imagery, but larger openings and overhanging geometries---tunnels, bridges---remain challenging. Materials follow ITU-R P.2040 permittivity and conductivity; objects whose electromagnetic behaviour these classes do not capture are unmodelled, as is surface roughness. Finally, KODAMA models outdoor environments only. Indoor scenes are comparatively well served by vision-based methods~\cite{ying2026horama, an2026tamingvisionpriorsdata, bejerano2026radartwinscenespecificmmwaveradar, yao2026rfdt}, yet a large fraction of mobile traffic originates indoors, making outdoor-indoor convergence a natural extension.

Our reconstruction stages are evaluated here only in combination, so a full factorial ablation isolating each component's contribution is our immediate next step. Longer term, because the propagation model is differentiable, measurements need not serve only as benchmarks but could act as a training signal, closing the loop between scene generation and channel prediction.
\section{Conclusions}
We demonstrate that site-specific RF digital twins can be reconstructed automatically at city scale from widely-available geospatial data, without site visits, bespoke capture, or calibration. Across three environments and frequencies from \qty{3.6}{\GHz} to \qty{28}{\GHz}, KODAMA's uncalibrated predictions improve on automated pipelines in current use by up to \qty{5.35}{\dB} point-to-point RMSE and come within \qty{0.22}{\dB} of a hand-built twin calibrated against measurement data. Beyond wireless, RF propagation offers 3D reconstruction an unusually strict and falsifiable test: scenes are judged not by how they look, but by whether signals propagate through them as they do at the real site.
{
    \small
    \bibliographystyle{ieeenat_fullname}
    \bibliography{main.bib}
}
\end{document}